\documentclass[runningheads]{llncs}

\usepackage{accv}

\usepackage{accvabbrv}

\usepackage{graphicx}
\usepackage{booktabs}

\usepackage[accsupp]{axessibility}  

\usepackage{hyperref}

\usepackage{orcidlink}
\usepackage{caption}
\usepackage{wrapfig}
\begin{document}

\title{Temporal Residual Bottleneck for Robust Asynchronous Collaborative Perception}

\titlerunning{Temporal Residual Bottleneck for Asynchronous Perception}

\author{
Melih Yazgan\inst{1,2} \and
Ahmed Abouelazm\inst{1,2} \and
J. Marius Zöllner\inst{1,2}
}

\authorrunning{M. Yazgan et al.}

\institute{
FZI Research Center for Information Technology\\
\email{surname@fzi.de}
\and
Karlsruhe Institute of Technology\\
\email{name.surname@kit.de}
}

\maketitle

\begin{abstract}
Collaborative perception extends the sensing range of autonomous vehicles, but its performance degrades when shared features arrive stale or incomplete.
Most latency-robust methods compensate delayed collaborator features through flow-guided alignment or direct feature transport.
In this work, we formulate asynchronous collaborative perception as temporal residual prediction.
Our Temporal Residual Bottleneck keeps a deterministic pose-warped collaborator feature as a conservative anchor and uses a $\Delta t$-conditioned xLSTM to extract residual temporal evidence from the available history.
A detector-facing residual bottleneck then applies only gated, regularized corrections before ego-side fusion, reducing the risk of overwriting reliable static structure when temporal correspondence is uncertain.
Experiments on DAIR-V2X and OPV2V show that our method is especially effective under severe fixed/irregular delays and packet drops. On DAIR-V2X, the reported checkpoint trades a small amount of synchronized peak accuracy for better robustness under stronger communication degradation.
Controlled diagnostics further indicate that direct feature transport has oracle headroom but can become unreliable when deployed without accurate correspondence.
These results support temporal residual fusion as a practical alternative for
asynchronous and incomplete collaborative perception. Code will be publicly
released at \url{https://url.fzi.de/8dk38}.
\end{abstract}

\section{Introduction}

Collaborative Perception (CP) extends the sensing range of connected autonomous
vehicles by aggregating complementary observations from nearby agents, which is
particularly useful under occlusion and other challenging sensing conditions
\cite{liu2023vehicletoeverythingautonomousdrivingsurvey,literature_review_perception_vs_cp,
yazgan2024collaborative,Han_2023,zhang2025e2e,missing_piece_fad}.
Recent work has increasingly considered practical constraints such as limited
bandwidth, adverse weather, heterogeneous sensors, and real-world deployment
\cite{yazgan2024realworld,yazgan2025slimcomm,yazgan2026radar,
zhou2026pragmatic,quantv2x,v2xrealo}. However, V2X communication is neither
instantaneous nor perfectly reliable: latency, asynchrony, and packet loss make
collaborator information stale or incomplete, degrading feature alignment and
fusion~\cite{syncnet,cobevflow,ffnet,lrcp,v2vlc,icav2x,codyntrust}.

Most intermediate-fusion methods exchange Bird's-Eye View (BEV) features and
assume that received collaborator information corresponds to the ego vehicle's
current scene state
\cite{chen2019fcooperfeaturebasedcooperative,opv2v,xu2022cobevtcooperativebirdseye,
hu2022where2commcommunicationefficientcollaborativeperception,yazgan2026dense}.
Under delay, however, moving objects remain temporally inconsistent even after
geometric alignment. Existing latency-aware approaches compensate stale
features through temporal synchronization, feature-flow prediction, or direct
feature transport~\cite{syncnet,cobevflow,ffnet,lrcp,codyntrust}. Such
compensation can become unreliable when motion or correspondence estimates are
inaccurate, motivating a more conservative interface to temporal evidence.

We therefore formulate asynchronous collaborative perception as
\emph{temporal residual prediction}. Our \emph{Temporal Residual Bottleneck}
preserves a deterministic pose-warped collaborator feature as an anchor and
learns only a gated residual correction. A $\Delta t$-conditioned temporal
branch, instantiated with xLSTM~\cite{beck2024xlstmextendedlongshortterm},
summarizes irregular BEV histories, while the residual bottleneck restricts how
this evidence modifies the feature passed to collaborative fusion. This avoids
replacing reliable geometric structure with an uncertain transported estimate.

Experiments on DAIR-V2X
\cite{yu2022dairv2xlargescaledatasetvehicleinfrastructure} and
OPV2V~\cite{opv2v} show improved robustness under severe delay, packet loss,
and pose noise. Controlled diagnostics further show that accurate oracle
transport has useful headroom, whereas deployable correspondence-based
transport can become brittle when applied directly in feature space.

Our contributions are threefold:
\begin{itemize}
    \item We propose a Temporal Residual Bottleneck for asynchronous
    collaborative perception that anchors stale collaborator features through
    deterministic pose warping and applies temporal evidence as a gated
    residual correction.

    \item We introduce a $\Delta t$-conditioned temporal formulation for
    irregular collaborator histories, instantiated with xLSTM, that preserves
 the pose-warped baseline instead of reconstructing the complete current feature.

    \item We provide controlled ablations and transport diagnostics that
    separate the effects of generic temporal prediction, the residual
    bottleneck, and correspondence-based transport, together with robustness
    evaluation on DAIR-V2X and OPV2V under delay, packet loss, and pose noise.
\end{itemize}
\section{Related Work}

\paragraph{Collaborative perception.}
Prior collaborative perception methods are commonly grouped into early, late,
and intermediate fusion. Early fusion aggregates raw observations before detection,
whereas late fusion combines object-level outputs from independently processed
agents~\cite{arnold2020cooperative,justo2025lff}. Intermediate fusion exchanges
task-relevant features and offers a practical trade-off between communication cost
and perception accuracy~\cite{hu2022where2commcommunicationefficientcollaborativeperception,
chen2026whispernet,xu_cosdh_2025,infocom}. Benchmarks and frameworks such as
OpenCDA, OPV2V, and DAIR-V2X have standardized multi-agent 3D perception
evaluation~\cite{opencda,opv2v,yu2022dairv2xlargescaledatasetvehicleinfrastructure}.
Our work follows this intermediate-fusion setting, but focuses on temporal
mismatch between the ego reference time and delayed collaborator features.

\paragraph{Asynchrony, latency, and concurrent architectures.}
A growing body of work addresses the mismatch between ego and collaborator timestamps.
SyncNet~\cite{syncnet} uses recurrent temporal modeling to estimate synchronized features from delayed histories, while CoBEVFlow~\cite{cobevflow} and FFNet~\cite{ffnet} utilize flow-based prediction or feature-warping mechanisms to align asynchronous features.
LRCP~\cite{lrcp} integrates latency-aware corrections into deformable attention, and CoDynTrust~\cite{codyntrust} builds on CoBEVFlow-style dense flow prediction with trust-based filtering to suppress unreliable compensated regions.
More recently, DelAwareCol$^\dagger$~\cite{DelAwareCol} introduces dilated temporal convolutions and multi-head temporal self-attention with relative delay encoding, while CATNet$^\dagger$~\cite{catnet}\footnote{$^\dagger$Lacks official open-source code for standardized benchmark evaluation.} addresses communication asynchrony and feature inconsistency through spatio-temporal recurrent synchronization, wavelet-based denoising, and adaptive feature selection.
While these methods demonstrate the necessity of modeling temporal misalignment, many rely on dense spatial attention, dense feature-flow prediction, or direct feature replacement.
In contrast, our method keeps a deterministic pose-warped baseline as the detector-facing anchor and learns only gated residual corrections from temporal evidence.

\paragraph{Robustness to spatial misalignment and pose errors.}
Realistic V2X deployments must handle not only temporal delay but also spatial misalignment from localization noise. 
CoDynTrust~\cite{codyntrust} builds on the dense flow prediction introduced by CoBEVFlow, but replaces direct flow-based compensation with linear extrapolation of Region-of-Interest (ROI) boxes and a Dynamic Feature Trust Module (DFTM) to suppress regions where motion compensation is unreliable. 
This trust-based masking improves robustness to noisy alignment, but it mainly filters stale evidence rather than explicitly predicting the unobserved current dynamic state. 
Our diagnostics on explicit object transport lead to a complementary conclusion: causal detector-derived box tracking is often too noisy to safely guide feature-space alignment, motivating our gated residual correction instead of direct transported-feature injection.

\paragraph{Position of our work.}
Our work intersects recent efforts in temporal delay compensation and spatial
robustness, but takes a different architectural approach. Existing methods
typically recover a synchronized collaborator representation through feature
prediction, spatial transport, or flow-based alignment before fusion. Instead,
we frame asynchronous collaborative perception as a
\emph{temporal residual prediction problem}. The proposed Temporal Residual
Bottleneck preserves a deterministic pose-warped collaborator feature as a
geometric anchor and lets learned temporal evidence modify it only through a
gated residual path, avoiding reconstruction or replacement of the complete
current collaborator feature.

Although the temporal history is processed per collaborator before fusion, the
formulation differs from conventional single-agent temporal feature prediction:
the learned branch estimates only residual evidence relative to the pose-warped
baseline rather than a complete current-time BEV representation. We instantiate
the temporal encoder with a $\Delta t$-conditioned xLSTM, but the residual
formulation is not tied to this architecture, as shown by the temporal-backbone
ablation in Sec.~\ref{sec:ablation}. Unlike 4D object tracking, our method does
not maintain persistent object identities or explicitly estimate trajectories;
it operates directly on dense collaborator BEV features before collaborative
fusion. This separation between geometric anchoring and learned temporal
correction is particularly useful under severe delay, missing evidence, or
spatial perturbations.
\section{Method}
\label{sec:method}
\begin{figure*}[h!]
    \centering
    \includegraphics[
    width=\textwidth,
    height=0.3\textheight,
    keepaspectratio
]{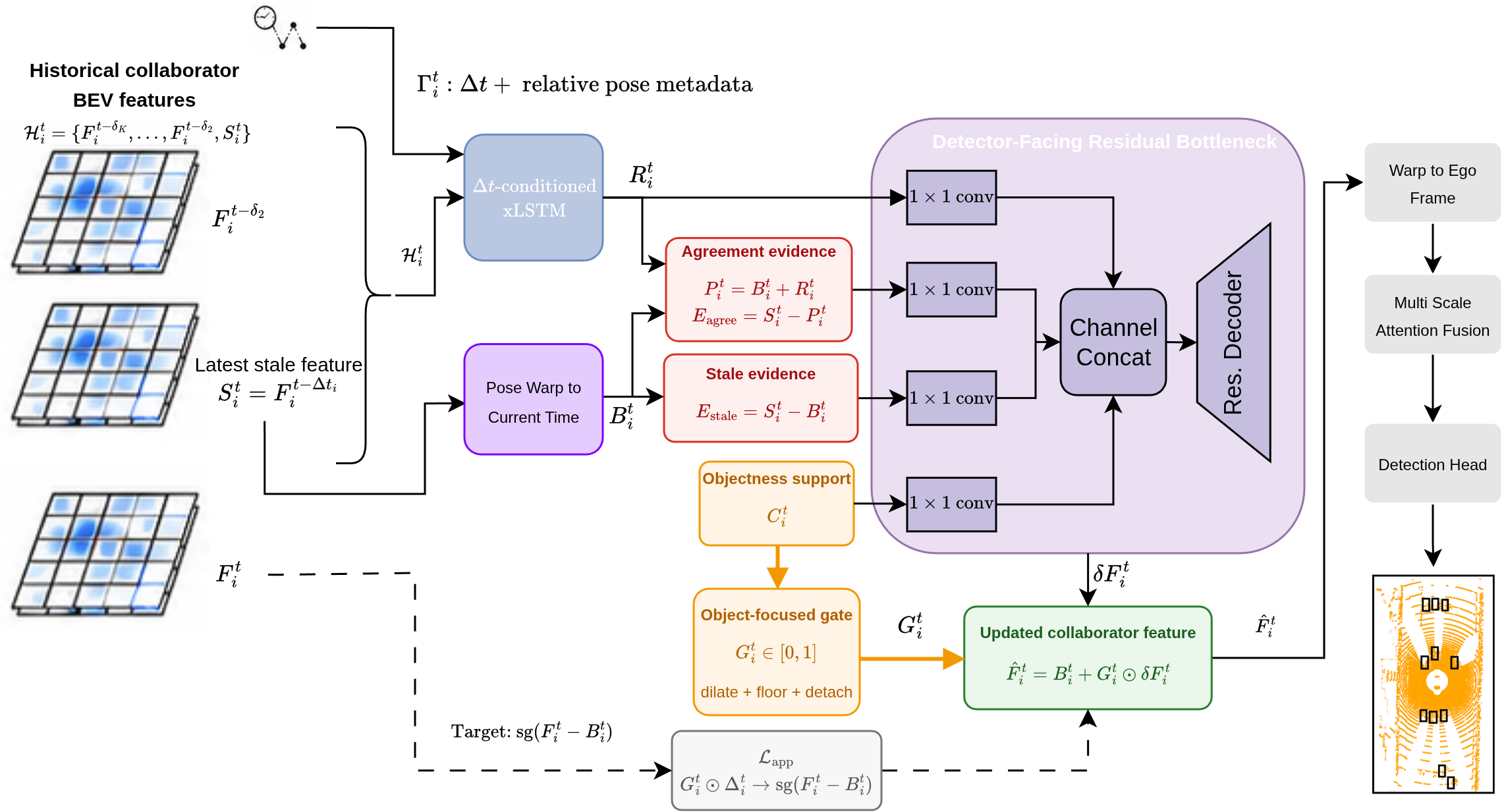}
    \caption{Overview of the proposed temporal residual bottleneck.
    Delayed collaborator BEV features are processed in the collaborator local frame.
    The latest stale feature $S_i^t$ is pose-warped into a geometric baseline $B_i^t$, while a $\Delta t$-conditioned xLSTM extracts temporal residual evidence $R_i^t$ from the available history.
    A detector-facing residual bottleneck predicts a correction $\delta F_i^t$, which is applied through an object-focused gate $G_i^t$ as
    $\hat{F}_i^t = B_i^t + G_i^t \odot \delta F_i^t$.
    The updated collaborator features are then warped to the ego frame and fused as multi-scale intermediate features for detection.}
    \label{fig:method_overview}
\end{figure*}
Fig.~\ref{fig:method_overview} illustrates the proposed Temporal Residual Bottleneck for asynchronous collaborative perception.
Given delayed collaborator BEV features, our method first performs temporal compensation in the collaborator local frame before ego-frame fusion.
It constructs a deterministic pose-warped geometric baseline, extracts temporal residual evidence from the available history using a $\Delta t$-conditioned xLSTM, and predicts a gated detector-facing correction.
The updated collaborator features are then transformed into the ego frame, fused as multi-scale intermediate features, and passed to the detection head.
\subsection{Problem Setup}

We consider intermediate-fusion collaborative perception with one ego vehicle
and one or more collaborators. Let $F_i^t \in \mathbb{R}^{C\times H\times W}$
denote the collaborator feature of agent $i$ at ego reference time $t$.
Under asynchronous communication, the latest available collaborator feature is
stale and is denoted by $S_i^t = F_i^{t-\Delta t_i}$, where
$\Delta t_i \geq 0$ is its realized age.

The available collaborator history may be irregularly spaced or partially
missing. We denote the causal history by
\[
\mathcal{H}_i^t =
\{(F_i^{t-\delta_k}, \delta_k, m_{i,k})\}_{k=1}^{K},
\]
where $\delta_k$ is the temporal offset and
$m_{i,k}\in\{0,1\}$ indicates whether the corresponding historical feature is
available. Received features may additionally be spatially incomplete under
packet loss.

\begin{wrapfigure}{r}{0.58\textwidth}
    \vspace{-8pt}
    \centering
    \includegraphics[
        width=\linewidth,
        trim={12pt 8pt 12pt 8pt},
        clip
    ]{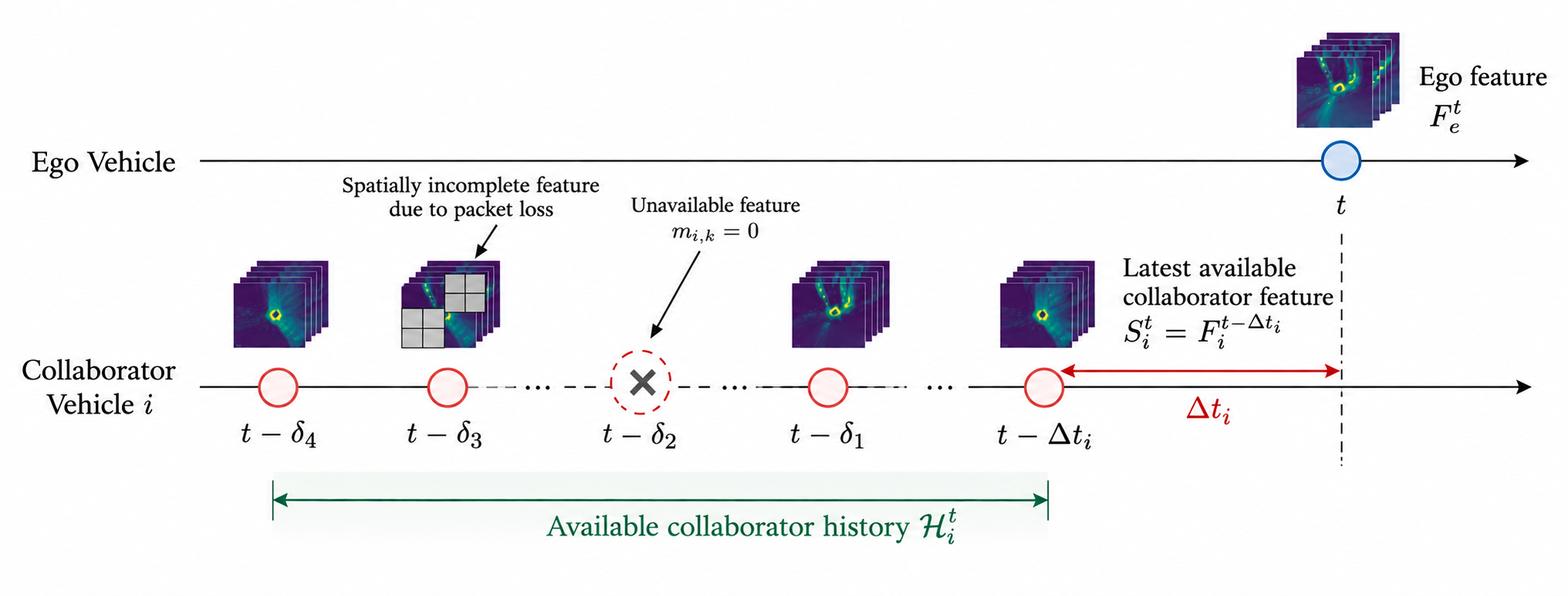}
    \caption{Asynchronous collaborative-perception setting. At ego time $t$,
    the latest collaborator feature $S_i^t$ has age $\Delta t_i$.
    Earlier features form the irregular history $\mathcal{H}_i^t$;
    features may be unavailable ($m_{i,k}=0$) or spatially incomplete.}
    \label{fig:delay_model}
    \vspace{-8pt}
\end{wrapfigure}

The method does not require prediction of future delay. Once a collaborator
feature arrives, its timestamp determines its realized age,
$\Delta t_i=t-t_{\mathrm{capture},i}$, which directly conditions the temporal
branch. The same checkpoint therefore handles different realized delays
without delay-regime prediction or model switching.

The goal is to construct a collaborator representation
$\hat{F}_i^t$ that supports current-time ego detection.
Rather than reconstructing the complete current collaborator feature $F_i^t$,
our deployable model retains a deterministic pose-warped baseline and learns
only a gated temporal residual correction before collaborative fusion.
\subsection{Pose-Warped Geometric Baseline}

The first component is a deterministic geometric baseline. The most recent valid collaborator feature $S_i^t$ is warped from its capture pose to the current collaborator reference time using the known relative pose transformation:
\[
B_i^t =
\mathcal{W}\!\left(S_i^t, T_{i,t-\Delta t_i \rightarrow i,t}\right),
\]
where $\mathcal{W}(\cdot)$ is the BEV feature warp and $T_{i,t-\Delta t_i \rightarrow i,t}$ maps the stale collaborator coordinate frame to the collaborator's current-time frame. This baseline is reliable for static scene structure and for ego/collaborator motion compensation, but it cannot fully correct independently moving objects.
The warp uses the same localization metadata assumed by standard V2X feature alignment: the stale packet is associated with the collaborator pose at its capture time, and, following the benchmark protocol used by prior asynchronous CP baselines,
the collaborator pose at the ego reference time is taken from the timestamped
localization stream. This is not an object-motion oracle; it compensates only known ego/collaborator pose changes. We evaluate sensitivity to imperfect pose metadata in Table~\ref{tab:pose_noise_robustness}.
After warping the latest stale feature into the current collaborator frame, we denote the resulting pose-warped geometric baseline as \(B_i^t\). This baseline
compensates known ego/collaborator pose changes, but does not model independent
object motion.

We therefore do not ask the network to synthesize the whole collaborator feature. Instead, the learned branch estimates temporal evidence relative to $B_i^t$, and a separate residual bottleneck decides how much of that evidence should be applied to the detector-facing feature. The design question is how to exploit temporal evidence without corrupting the baseline when the evidence is uncertain.



\subsection{Time-Conditioned Temporal Evidence}

We use a recurrent temporal predictor to summarize the collaborator history:
\[
R_i^t = \Psi_\theta(\mathcal{H}_i^t, \Gamma_i^t),
\]
where $\Psi_\theta$ is an xLSTM-based predictor and $\Gamma_i^t$ contains
temporal offsets and relative pose metadata. BEV cells are processed as
temporal tokens so that each location's short history is summarized by
recurrent memory.

Delay is encoded using fixed multi-frequency Fourier features
\cite{mildenhall2020nerf,tancik2020fourier},
\[
\phi(\Delta t)=
\left[
\Delta t,\,
\{\sin(2^k\pi\Delta t),\cos(2^k\pi\Delta t)\}_{k=0}^{4}
\right],
\]
followed by a learned FiLM conditioner. A separate pose/time token adds
relative translation and periodic yaw information.

The recurrent output $R_i^t$ is treated as temporal residual evidence rather
than a standalone transported feature and is therefore routed through the
residual bottleneck.

\subsection{Detector-Facing Residual Bottleneck}
The final model uses an attention-free residual bottleneck. We refer to the
resulting collaborator representation as \emph{detector-facing} because it is
the feature directly exposed to the subsequent collaborative fusion and
detection stages. The bottleneck converts recurrent temporal evidence into a
residual correction while preserving the pose-warped baseline as the default
prediction.

Let \(S_i^t\) be the latest stale feature and \(B_i^t\) its pose-warped
geometric baseline. The residual bottleneck also receives the stale feature
itself, so that it can compare the uncorrected stale evidence against the
pose-warped baseline and the recurrent prediction.
\[
P_i^t = B_i^t + R_i^t
\]
is the raw recurrent residual prediction. The residual bottleneck receives three evidence terms and an object-support confidence map:
\[
E_{\mathrm{rec}} = R_i^t, \qquad
E_{\mathrm{stale}} = S_i^t - B_i^t, \qquad
E_{\mathrm{agree}} = S_i^t - P_i^t.
\]
Here $E_{\mathrm{stale}}$ is not treated as a physical current-frame residual; it is a warp-discrepancy cue computed in the shared BEV tensor layout to indicate how much deterministic pose compensation changes the latest stale evidence. The agreement term exposes whether the recurrent prediction is consistent with the most recent collaborator evidence.

Each evidence term is projected into a compact bottleneck space. With a confidence map $C_i^t$ and learned projections $\pi_{\cdot}$, the correction is
\[
\delta F_i^t =
s \cdot
\Omega_\phi\!\left(
\left[
\pi_{\mathrm{rec}}(E_{\mathrm{rec}}),
\pi_{\mathrm{stale}}(E_{\mathrm{stale}}),
\pi_{\mathrm{agree}}(E_{\mathrm{agree}}),
\pi_c(C_i^t)
\right]
\right),
\]
where $\Omega_\phi$ is a small convolutional residual decoder and $s$ is a residual scale. The decoder is zero-initialized, so the model starts from the conservative baseline $B_i^t$ and learns only corrections that improve detection.

To avoid applying residuals uniformly over the BEV map, the model uses an object-focused gate $G_i^t\in[0,1]^{1\times H\times W}$. The gate is derived from detector objectness maps, not from ground-truth labels or oracle boxes. In our implementation, these maps are obtained by applying the lightweight Probability Score Map (PSM)/classification head to the same cached collaborator BEV features already available to the temporal module; therefore, the gate does not require transmitting an additional dense objectness tensor. Concretely, we collapse the resulting collaborator PSM logits into a single objectness map, take the most recent stale objectness and its pose-warped version, and use their union as object support. This support is dilated, assigned a small floor value, and detached before modulating the residual. The stale-and-warped union makes the gate less brittle under delay: stale objectness preserves evidence before warping, while the warped objectness covers the estimated current position. Ground-truth labels are used only to define auxiliary foreground masks during training; they are not used to route residuals at inference time. The final collaborator feature is
\[
\hat{F}_i^t = B_i^t + G_i^t \odot \delta F_i^t.
\]
This is the key detector-facing interface: temporal evidence can modify the baseline, but only through a learned, gated residual path. The scalar $s$ controls the residual amplitude, the decoder is zero-initialized, and the training objective below regularizes over-applied corrections. We do not claim a formal hard bound such as spectral normalization or per-cell clipping; the safety argument is architectural and empirical, not a mathematical guarantee. The method therefore avoids direct feature overwrite by uncertain spatial transport.

\subsection{Fusion and Detection}

After temporal compensation, all collaborator features are aligned into the ego frame and processed by a Where2Comm-style multi-scale intermediate-fusion backbone with confidence-based communication masking and the same classification and box-regression heads~\cite{hu2022where2commcommunicationefficientcollaborativeperception}; the temporal residual module is inserted before this ego-frame fusion stage:
\[
F_{\mathrm{fused}}^t =
\Phi\!\left(F_e^t, \hat{F}_1^t,\ldots,\hat{F}_{N-1}^t\right).
\]
This keeps the detector architecture fixed; the proposed contribution is the temporal residual interface that prepares collaborator features before fusion.


\subsection{Training Objective}

The main objective is the standard detection loss,
\[
\mathcal{L}_{\mathrm{det}}
=
\mathcal{L}_{\mathrm{cls}}
+
\lambda_{\mathrm{box}}\mathcal{L}_{\mathrm{box}}.
\]
We do not use a feature reconstruction loss in the final model. Instead, we add a small detector-facing applied-residual objective on foreground or label-supported regions. Let
\[
D_i^{t,*}=\mathrm{sg}(F_i^t-B_i^t)
\]
be the synchronized residual target available during training, where $\mathrm{sg}(\cdot)$ denotes stop-gradient. Thus, the synchronized current feature supervises the correction but is not used as an inference input. Let $A_i^t=G_i^t\odot\delta F_i^t$ be the residual actually applied to the detector-facing feature, and let $M_i^t$ denote the foreground/label mask. The applied-residual objective is
\[
\mathcal{L}_{\mathrm{app}}
=
\sum_i
\left(
w_{\mathrm{dir}}\mathcal{L}_{\mathrm{dir}}^i
+w_{\mathrm{mag}}\mathcal{L}_{\mathrm{mag}}^i
+w_{\mathrm{imp}}\mathcal{L}_{\mathrm{imp}}^i
+w_{\mathrm{over}}\mathcal{L}_{\mathrm{over}}^i
\right),
\]
where
\[
\mathcal{L}_{\mathrm{dir}}^i =
\left\langle
1-\cos(A_i^t,D_i^{t,*})
\right\rangle_{M_i^t},
\qquad
\mathcal{L}_{\mathrm{mag}}^i =
\left\langle
\mathrm{smooth}\text{-}\ell_1
\left(
\|A_i^t\|_2,\|D_i^{t,*}\|_2
\right)
\right\rangle_{M_i^t}.
\]
The remaining two terms discourage harmful corrections:
\[
\mathcal{L}_{\mathrm{imp}}^i =
\left\langle
\mathrm{ReLU}
\left(
e(B_i^t+A_i^t,F_i^t)-e(B_i^t,F_i^t)+m
\right)
\right\rangle_{M_i^t},
\]
\[
\mathcal{L}_{\mathrm{over}}^i =
\left\langle
\mathrm{ReLU}
\left(
\|A_i^t\|_2-\rho\|D_i^{t,*}\|_2
\right)
\right\rangle_{M_i^t}.
\]
Here $\langle\cdot\rangle_{M_i^t}$ denotes the average over masked BEV cells, $\cos(\cdot,\cdot)$ uses $\epsilon$-stabilized normalization, $e(\cdot,\cdot)$ is the channel-mean absolute feature error, $m$ is a small improvement margin, and $\rho$ controls the allowed residual-to-target magnitude ratio. The numerical weights and margins are listed in the Supp.~A.
The total training loss is
\[
\mathcal{L}
=
\mathcal{L}_{\mathrm{det}}
+
\lambda_{\mathrm{app}}\mathcal{L}_{\mathrm{app}},
\]
This objective supervises the correction actually exposed to the detector rather than reconstructing full BEV features. The model is trained end-to-end with the standard detection objective, so the residual interface is optimized jointly with the downstream fusion and detection losses.

\section{Experiments}
\label{sec:experiments}
\subsection{Experimental Setup}

We evaluate on DAIR-V2X~\cite{yu2022dairv2xlargescaledatasetvehicleinfrastructure} and OPV2V Culver City~\cite{opv2v} in the LiDAR-only intermediate-fusion setting using single-class vehicle detection. For DAIR-V2X, all methods use the same BEV discretization with voxel size $(0.4,0.4,4.0)$, a LiDAR range of $201.6\,\mathrm{m}\times80\,\mathrm{m}$, raw BEV size $(200,504)$, and stride-2 detection grid $(100,252)$. For OPV2V, we use a LiDAR range of $280\,\mathrm{m}\times80\,\mathrm{m}$. All methods in Table~\ref{tab:dair_baselines} and \ref{tab:opv2v_culver} use PointPillars  as the LiDAR backbone~\cite{lang2019pointpillarsfastencodersobject}. We report AP at IoU thresholds 0.3, 0.5, and 0.7, and use AP@0.7 as the primary metric.

We evaluate seven communication settings: synchronized input, fixed and irregular 100/300/500\,ms average delay. Fixed delay uses a deterministic first history offset, while irregular delay samples temporal offsets from a binomial history process with the same expected delay. We also evaluate a joint delay-plus-packet-drop stress setting, where contiguous regions of non-ego collaborator BEV features are removed in burst mode at inference time.

\paragraph{Fair comparison protocol.}
All methods are evaluated with the same validation split, BEV grid, detection
post-processing, delay settings, and AP computation. Published reference numbers
are reported where available, while reproduced in-framework integrations are
marked separately. Each reproduced baseline follows its official training protocol.
Packet drop is activated only at inference time and applied consistently to all
compared methods.

Our method additionally consumes a four-frame collaborator history for temporal
residual prediction. This history is obtained from an ego-side cache of packets
received in previous timesteps; collaborators do not resend the last $K$ frames at
every step. Thus, the current communication payload is unchanged, while
receiver-side memory and computation increase with history length. All
latency-aware baselines are evaluated with their official temporal-cache or
history settings, while our method uses $k=4$ cached collaborator frames; thus,
the comparison follows each method's intended asynchronous input protocol rather
than enforcing a single-frame setting. We report history-length ablations
separately.
\subsection{Implementation Details}
We train the model end-to-end with Adam using learning rate $2\times10^{-3}$,
weight decay $10^{-4}$, $\epsilon=10^{-10}$, cosine decay, and a 7-epoch
warmup from $2\times10^{-5}$. DAIR-V2X models are trained for 60 epochs
with batch size 2, while OPV2V models are trained for 30 epochs with batch
size 1. Random asynchronous training samples binomial delays with
$p\in[0.05,0.30]$, and the reported checkpoint is selected by validation
performance. The temporal branch uses four history sweeps ($k=4$), embedding
dimension 64, and an m-s-m xLSTM stack. The residual bottleneck uses a
64-channel bottleneck, residual scale 1.5, zero-initialized output, and
applied-residual loss weight 0.02. Packet drop is activated only at inference and is used as a severe
missing-feature stress test rather than a calibrated wireless-channel model.
Further implementation details are provided in Supp.~A, and runtime is analyzed
in Supp.~D.
\subsection{Comparison to DAIR-V2X Baselines}

Table~\ref{tab:dair_baselines} compares our method against DAIR-V2X baselines under synchronized input and two irregular-delay settings. 
\begin{table}[h!]
\centering
\caption{\textbf{DAIR-V2X baseline comparison.} AP@0.5/AP@0.7 under synchronized and irregular-delay settings. ``pub.'' denotes published reference numbers from CoDynTrust~\cite{codyntrust}; ``rep.'' denotes reproduced integrations in our framework. \textbf{Bold} and \underline{underline} denote the best and second-best results, respectively.}
\label{tab:dair_baselines}
\small
\setlength{\tabcolsep}{3.5pt}
\begin{tabular}{lccc}
\toprule
Method & Sync & Irr.\ 300\,ms & Irr.\ 500\,ms \\
\midrule
Where2Comm+SyncNet~\cite{syncnet} (pub.) & 0.792 / \underline{0.692} & 0.732 / 0.625 & 0.710 / 0.612 \\
CoBEVFlow~\cite{cobevflow} (pub.) & 0.775 / 0.663 & 0.742 / 0.626 & 0.732 / 0.621 \\
CoDynTrust~\cite{codyntrust} (pub.) & 0.792 / \textbf{0.700} & 0.743 / \textbf{0.643} & \underline{0.736} / \textbf{0.637} \\
\midrule
Where2Comm+SyncNet (rep.) & 0.782 / 0.688 & 0.738 / 0.606 & 0.732 / 0.603 \\
CoBEVFlow (rep.) & 0.788 / 0.660 & 0.739 / 0.597 & 0.718 / 0.586 \\
CoDynTrust (rep.) & \textbf{0.798} / 0.668 & 0.726 / 0.586 & 0.719 / 0.582 \\
LRCP & \underline{0.795} / 0.669 & \textbf{0.769} / 0.639 & 0.735 / 0.611 \\
Ours & 0.781 / 0.664 & \underline{0.763} / \underline{0.640} & \textbf{0.753} / \underline{0.635} \\
\bottomrule
\end{tabular}
\end{table}

Because our implementation builds on the CoDynTrust codebase, we report both published reference numbers and reproduced in-framework integrations. The reproduced rows provide the closest implementation-controlled comparison, while the published rows indicate the expected performance of the original methods. A CoBEVFlow ROI-mask diagnostic is provided in Supp.~E; Table~\ref{tab:dair_baselines} keeps the released code path for reproducibility. We mark the best result in bold and the second-best result with underlining for each AP metric and scenario.

Table~\ref{tab:dair_baselines} shows a robustness trade-off rather than a broad synchronized-SOTA result. In the synchronized setting, published CoDynTrust and W2C+SyncNet remain stronger at AP@0.7, and reproduced CoDynTrust is strongest at AP@0.5. Under irregular delay, however, our method becomes increasingly competitive: it is near the best AP@0.7 at 300\,ms and achieves the best AP@0.5 at 500\,ms, while remaining within a small margin of the published CoDynTrust AP@0.7 reference. In particular, LRCP drops from 0.669 to 0.611 AP@0.7 from synchronized input to irregular 500\,ms delay ($-0.058$), whereas our method drops from 0.664 to 0.635 ($-0.029$). This supports the intended claim that the residual temporal interface preserves localization under severe asynchrony, rather than maximizing synchronized peak accuracy. Additional mixed synchronized/asynchronous fine-tuning experiments in the Supp.~F show that the small synchronized gap is not intrinsic to the
architecture.

\paragraph{Communication footprint.}
\begin{wraptable}{r}{0.50\textwidth}
\centering
\caption{\textbf{Communication footprint} under DAIR-V2X irregular 300\,ms delay.
Payload is estimated per collaborator BEV feature packet.}
\label{tab:comm_payload}
\resizebox{\linewidth}{!}{%
\begin{tabular}{lcccc}
\toprule
Method & AP@0.7 & \texttt{Comm. Rate} & Payload \\
       &        &              ($r$)         & (MB)  \\
\midrule
Where2Comm+SyncNet (rep.) & 0.606 & 0.1225 & 0.79 \\
CoBEVFlow (rep.)          & 0.597 & 0.1441 & 0.93 \\
CoDynTrust (rep.)         & 0.586 & 0.2278 & 1.47 \\
LRCP                      & 0.639 & 1.0000 & 6.45 \\
Ours                      & 0.640 & 0.1572 & 1.01\\
\bottomrule
\end{tabular}%
}
\vspace{-1em}
\end{wraptable}
We estimate the transmitted BEV feature payload under DAIR-V2X irregular
300\,ms delay. All methods in Table~\ref{tab:comm_payload} except LRCP use
Where2Comm-style spatial communication masking, whereas LRCP transmits dense
BEV features. For sparse methods, the payload is computed as
$B=rHWD\cdot4$ bytes FP32 features, where $r$ is the logged communication rate~\cite{hu2022where2commcommunicationefficientcollaborativeperception}. The
communicated scale feature has shape $D\times H\times W=64\times100\times252$.
Thus, dense LRCP uses 6.45\,MB per collaborator packet, while ours uses
1.01\,MB with comparable AP@0.7. The estimate counts feature tensors only.
\subsection{OPV2V Culver City Results}
Table~\ref{tab:opv2v_culver} reports the OPV2V Culver City benchmark under synchronized input and irregular-delay settings. Unlike DAIR-V2X, the reproduced OPV2V baselines do not show a deficit relative to the original CoDynTrust-paper reference numbers; for example, the original paper reports 0.792 AP@0.5 for Where2Comm, Where2Comm+SyncNet, and CoDynTrust, while our reproduced synchronized baselines are comparable or higher. 
\begin{table}[h!]
\centering
\caption{\textbf{OPV2V Culver City comparison.} AP@0.5/AP@0.7 under synchronized and irregular-delay settings.}
\label{tab:opv2v_culver}
\small
\setlength{\tabcolsep}{3.5pt}
\begin{tabular}{lccc}
\toprule
Method & Sync & Irr.\ 300\,ms & Irr.\ 500\,ms \\
\midrule
Where2Comm~\cite{hu2022where2commcommunicationefficientcollaborativeperception} & 0.870 / 0.786 & 0.801 / 0.664 & 0.768 / 0.639 \\
Where2Comm+SyncNet~\cite{syncnet} & 0.870 / 0.786 & 0.811 / 0.706 & 0.789 / 0.691 \\
CoBEVFlow~\cite{cobevflow} & 0.837 / 0.783 & 0.822 / 0.740 & 0.819 / 0.732 \\
CoDynTrust~\cite{codyntrust} & 0.865 / 0.819 & 0.857 / 0.795 & 0.853 / 0.783 \\
Ours & \textbf{0.895 / 0.835} & \textbf{0.884 / 0.818} & \textbf{0.868 / 0.799} \\
\bottomrule
\end{tabular}
\end{table}
Thus, the OPV2V comparison is not driven by weakened reproduced baselines. In this setting, our residual temporal model is the strongest method in all reported scenarios. It improves both AP@0.5 and AP@0.7 under both synchronized and irregular 300/500\, ms-delay inputs, including over the strongest listed CoDynTrust baseline. This difference is consistent with the benchmark structure: OPV2V provides simulated multi-CAV V2V sequences with cleaner pose/time consistency and richer temporal redundancy, enabling cached history to improve even when fusion is synchronized. DAIR-V2X is a real-world V2I system with only vehicle-infrastructure collaboration and noisier calibration/timing, making the benefit mainly apparent under severe delay, packet drop, and pose noise.


\subsection{Robustness to Delay and Packet Drop}

Table~\ref{tab:dair_delay_robustness_combined} compares our method with LRCP~\cite{lrcp} under clean delay and joint delay-plus-packet-drop stress. LRCP has higher synchronized and mild-delay AP, while our method preserves AP better as delay becomes severe. From synchronized input to irregular 500\,ms delay, LRCP drops by 0.058 AP@0.7, while ours drops by 0.029, reducing the degradation by 50\%. The result should therefore be read as a robustness trade-off rather than a broad DAIR-V2X SOTA claim: the residual model is comparable around 300\,ms and stronger at 500\,ms in both fixed and irregular settings. The packet-drop setting is intentionally severe and is not meant to approximate a calibrated V2X channel; instead, it acts as a worst-case proxy for bursty feature-packet interruptions that can arise under congested or degraded V2X communication. Further discussions and visualizations are provided in Supp.~B.
\begin{table*}[h!]
\centering
\caption{\textbf{DAIR-V2X robustness under delay and packet-drop stress.}
AP@0.7 across fixed and irregular delay scenarios. Positive $\Delta$ means our method is better.
Left: delay-only setting. Right: joint delay and packet-drop stress.}
\label{tab:dair_delay_robustness_combined}
\footnotesize 
\setlength{\tabcolsep}{2.5pt} 

\begin{minipage}[t]{0.48\textwidth}
\centering
\textbf{(a) Delay robustness}\\[2mm]
\begin{tabular}{lccc}
\toprule
Scenario & LRCP & Ours & $\Delta$ \\
\midrule
Sync & 0.669 & 0.664 & -0.005 \\
Fixed 100\,ms & 0.664 & 0.654 & -0.010 \\
Fixed 300\,ms & 0.652 & 0.640 & -0.011 \\
Fixed 500\,ms & 0.603 & 0.635 & +0.033 \\
Irregular 100\,ms & 0.663 & 0.654 & -0.009 \\
Irregular 300\,ms & 0.639 & 0.640 & +0.001 \\
Irregular 500\,ms & 0.611 & 0.635 & +0.023 \\
\bottomrule
\end{tabular}
\end{minipage}%
\hfill
\begin{minipage}[t]{0.50\textwidth} 
\centering
\textbf{(b) Delay + packet drop}\\[2mm]
\begin{tabular}{lccc}
\toprule
Scenario & LRCP & Ours & $\Delta$ \\
\midrule
Sync + drop & 0.655 & 0.652 & -0.003 \\
Fixed 100\,ms + drop & 0.648 & 0.643 & -0.005 \\
Fixed 300\,ms + drop & 0.628 & 0.629 & +0.001 \\
Fixed 500\,ms + drop & 0.598 & 0.622 & +0.025 \\
Irregular 100\,ms + drop & 0.650 & 0.641 & -0.009 \\
Irregular 300\,ms + drop & 0.622 & 0.629 & +0.007 \\
Irregular 500\,ms + drop & 0.604 & 0.621 & +0.017 \\
\bottomrule
\end{tabular}
\end{minipage}
\end{table*}

These results support the intended use case of the residual interface. When communication is synchronized or only mildly delayed, direct alignment can be highly effective. In the presence of larger delays or missing feature regions, however, a conservative residual correction preserves localization more reliably because it does not overwrite the detector-facing feature with an uncertain transported estimate.

\subsection{Robustness to Pose and Localization Error}

Following~\cite{cobevflow,codyntrust}, we further stress this setting by perturbing only the inference-time fusion poses, keeping detector checkpoints fixed. Translation and yaw perturbations are sampled from $\mathcal{N}(0,\sigma_t)$ and $\mathcal{N}(0,\sigma_\theta)$, with matched $\sigma_t \in \{0.1,0.3,0.5\}$\,m and $\sigma_\theta \in \{0.1,0.3,0.5\}^{\circ}$ in Table~\ref{tab:pose_noise_robustness}. The zero-noise case corresponds to the irregular 300\,ms setting in Table~\ref{tab:dair_baselines}. LRCP remains competitive at mild noise, but its AP@0.7 drops sharply as error increases. Our model better preserves strict localization at medium and high noise levels, utilizing temporal evidence as a conservative detector-facing correction rather than direct spatial overwrite. Qualitative results are provided in Supp.~C.
\begin{table}[h!]
\centering
\caption{\textbf{Robustness to pose/localization noise.} DAIR-V2X irregular 300\,ms delay with inference-time fusion-pose noise. The noise column gives matched $(\sigma_t,\sigma_\theta)$ in meters/degrees. We report AP@0.5/AP@0.7.}
\label{tab:pose_noise_robustness}
\small
\setlength{\tabcolsep}{4pt}
\begin{tabular}{lcc}
\toprule
Noise $(\sigma_t,\sigma_\theta)$ & LRCP & Ours  \\
\midrule
0.1\,m / 0.1$^\circ$ & 0.784 / 0.630 & 0.761 / 0.634 \\
0.3\,m / 0.3$^\circ$ & 0.722 / 0.560 & 0.738 / 0.602  \\
0.5\,m / 0.5$^\circ$ & 0.668 / 0.534 & 0.712 / 0.589  \\
\bottomrule
\end{tabular}
\end{table}

\subsection{Ablation and Diagnostic Findings}
\label{sec:ablation}
Table~\ref{tab:ablation_temporal} isolates the contributions of temporal
prediction and the detector-facing residual interface. The pose-warped
baseline alone achieves 0.622 AP@0.7. Using the residual bottleneck without
xLSTM evidence improves this to 0.628, while a conventional xLSTM that
directly predicts the current collaborator feature likewise reaches only
0.628. Conversely, retaining the xLSTM while removing the residual bottleneck
yields 0.630. The complete Temporal Residual Bottleneck reaches 0.640,
indicating that neither generic temporal prediction nor the residual interface
alone explains the improvement; their combination is required. Removing the
object-focused gate also reduces AP@0.7 to 0.630, supporting its role in
restricting temporal corrections to detector-supported regions while
preserving the pose-warped baseline elsewhere.

The transport diagnostics test whether explicit motion compensation should be
injected into detector-facing features. ``Oracle full residual'' applies the
synchronized residual target $F_i^t-B_i^t$ and reaches 0.660 AP@0.7, showing
the headroom of an ideal correction. ``Past-track teacher transport'' uses
annotation-derived correspondences and reaches 0.645, indicating that accurate
object transport can help. In contrast, ``Predicted transport applied'' uses
causal detector-derived correspondences and reaches only 0.631, showing that
deployable correspondence estimates are not reliable enough for direct feature
overwrite. Dense-flow-style and local cross-attention variants showed the same
general trend. We therefore retain the conservative residual bottleneck as the
final detector-facing interface.
\begin{table}[h!]
\centering

\begin{minipage}[t]{0.48\textwidth}
\centering
\caption{\textbf{DAIR-V2X ablation and diagnostics.}
All rows use the Irr.\ 300\,ms delay.}
\label{tab:ablation_temporal}
\resizebox{\linewidth}{!}{%
\begin{tabular}{lccc}
\toprule
Variant & Deploy. & AP@0.5 & AP@0.7 \\
\midrule
Pose-warped baseline only
    & Yes & 0.740 & 0.622 \\

Residual bottleneck w/o xLSTM evidence
    & Yes & 0.747 & 0.628 \\

Conventional xLSTM feature prediction
    & Yes & 0.742 & 0.628 \\

xLSTM w/o residual bottleneck
    & Yes & 0.748 & 0.630 \\

Residual bottleneck w/o object gate
    & Yes & 0.751 & 0.630 \\

\textbf{Temporal Residual Bottleneck (full)}
    & Yes & 0.765 & 0.640 \\

\midrule
Oracle full residual
    & No & 0.776 & 0.660 \\

Past-track teacher transport
    & No & 0.768 & 0.645 \\

Predicted transport applied
    & Yes & 0.751 & 0.631 \\
\bottomrule
\end{tabular}%
}
\end{minipage}\hfill
\begin{minipage}[t]{0.48\textwidth}
\centering
\caption{\textbf{Temporal backbone and history ablation.} AP@0.7 on DAIR-V2X, k is the historical frame number.}
\label{tab:xlstm_block_ablation}
\resizebox{\linewidth}{!}{%
\begin{tabular}{lccc}
\toprule
Configuration & Sync & Irr.\ 300\,ms & Irr.\ 500\,ms \\
\midrule
xLSTM-s, k=4 & 0.652 & 0.635 & 0.628  \\
xLSTM-m, k=4 &  0.643 & 0.626 & 0.618 \\
xLSTM-sms, k=4 & 0.645 & 0.625 & 0.617  \\
xLSTM-msm, k=2 &  0.642 & 0.620 & 0.613 \\
GRU, k=4 & 0.649 & 0.630 & 0.627 \\
xLSTM-msm, k=4 (Ours) & 0.664 & 0.640 & 0.635  \\
\bottomrule
\end{tabular}%
}
\end{minipage}

\end{table}
We also tested pose/time conditioning variants. Historical matched runs show a small AP@0.7 improvement from adding a pose/time token, but this effect is smaller than those of the xLSTM and residual bottleneck. This diagnostic motivates the final design choice: temporal evidence is useful,
but it should enter the detector through a gated residual path rather than
through direct feature overwrite. We therefore treat pose/time conditioning as a supporting component rather than the main source of performance.
\paragraph{Temporal backbone and history length.}
Table~\ref{tab:xlstm_block_ablation} varies the recurrent backbone,
xLSTM block sequence, and history length while keeping the temporal-residual
interface fixed. Replacing xLSTM with a three-layer GRU remains effective,
reaching 0.630 AP@0.7 at Irr.\ 300\,ms, which shows that the proposed
residual formulation is not specific to xLSTM. The final $m$-$s$-$m$
xLSTM configuration with four history frames ($k=4$) nevertheless performs
best across all three evaluated settings. Shorter history and alternative
xLSTM block patterns consistently underperform, indicating that both
sufficient temporal context and the selected mixed-memory configuration
contribute to the final result.

\section{Conclusion}
We presented a temporal residual bottleneck for asynchronous collaborative perception under delayed and incomplete communication. The method preserves a deterministic pose-warped collaborator feature as a conservative anchor and injects only gated residual corrections from a $\Delta t$-conditioned xLSTM. This conservative detector-facing interface avoids direct feature overwrite when temporal correspondence is uncertain. Experiments on DAIR-V2X show that the reported model does not maximize peak synchronous AP, but it degrades more gracefully than a flow-guided LRCP baseline under severe fixed delay, irregular delay, and joint packet-drop stress. Results on OPV2V Culver City further show that the same residual formulation can improve both synchronized and delayed performance. The current study still has limitations: the packet-drop benchmark is an aggressive stress test, and the residual gate is not a formal safety guarantee. Future work will study residual norm diagnostics and communication-aware residual selection under more realistic packet-loss processes.
\paragraph{\textbf{Acknowledgments.}}The research leading to these results is funded by the German Federal Ministry for Economic Affairs and Energy within the project “NXT GEN AI METHODS – Generative Methoden für Perzeption, Prädiktion und Planung". The authors would like to thank the consortium for the successful cooperation.

\clearpage
\setcounter{page}{1}
\appendix
\pagebreak
\section*{Supplementary Material for Temporal Residual Bottleneck for Robust Asynchronous Collaborative Perception}
\addcontentsline{toc}{section}{Appendix}
\section{Additional Implementation Details}
\label{sec:supp_impl}

Table~\ref{tab:supp_hyperparams} lists the implementation details omitted from the main paper for readability. These settings correspond to the DAIR-V2X and OPV2V temporal residual model reported in the main experiments.

\begin{table}[h!]
\centering
\caption{\textbf{Supplementary hyperparameters for scratch training.} Detailed training, temporal branch, residual bottleneck, object gate, and packet-drop settings.}
\label{tab:supp_hyperparams}
\small
\begin{tabular}{lp{0.75\linewidth}}
\toprule
Component & Setting \\
\midrule
\multicolumn{2}{l}{\textbf{Training Configuration}} \\
Optimizer & Adam, learning rate $2\times10^{-3}$, weight decay $10^{-4}$, $\epsilon=10^{-10}$ \\
Schedule & Cosine decay with 7-epoch warmup (warmup LR $2\times10^{-5}$); \textbf{DAIR-V2X:} 60 epochs \textbar{} \textbf{OPV2V:} 30 epochs \\
Batch size & \textbf{DAIR-V2X:} 2 \textbar{} \textbf{OPV2V:} 1 \\
Async training & Random binomial delay with $p\in[0.05,0.30]$ \\
\midrule
\multicolumn{2}{l}{\textbf{Architecture \& Temporal Branch Settings}} \\
History & Four history sweeps, so xLSTM context length 4 \\
xLSTM branch & Embedding dimension 64, $m$-$s$-$m$ block order, 4 mLSTM heads, dropout 0 \\
Conditioning & Pose/time token hidden dim 128, 4 Fourier bands, scale 0.1, zero-init head \\
Residual bottleneck & 64-ch bottleneck, 8 conf ch, hidden dim 256, kernel 3, scale 1.5, zero-init output \\
Object-focused gate & Stale-and-warped support union, dilation 4, floor 0.05, detached gradients \\
\midrule
\multicolumn{2}{l}{\textbf{Objective \& Stress Test Settings}} \\
Detector objective & No feature recon loss; applied-res weight $\lambda_{\mathrm{app}}=0.02$; $w_{\mathrm{dir}}=1.0$, $w_{\mathrm{mag}}=0.25$, $w_{\mathrm{imp}}=0.25$, $w_{\mathrm{over}}=0.1$; improvement margin $m=5\times10^{-5}$, over-application ratio $\rho=1.5$; FG/BG weight 3.0/1.0, label dilation 2 \\
Packet-drop test & Inference only; non-ego burst drop $p=1.0$, tile $8\times8$, burst len 12 (ego safe) \\
\bottomrule
\end{tabular}
\end{table}

\section{Contextualizing Packet Drop as Severe V2X Degradation}

In Sec.~4.5 of the main paper, we evaluate the Temporal Residual Bottleneck under a joint delay-plus-packet-drop stress test. This setting is not intended to be a calibrated wireless-channel simulator. Instead, it serves as a worst-case proxy for feature-packet interruptions under lossy or impaired V2X communication, a setting studied in recent collaborative perception work~\cite{v2vlc,icav2x}.

In practical feature-level collaborative perception, BEV tensors must be serialized, packetized, transmitted, and reconstructed at the receiver. When packets are delayed, dropped, or arrive incompletely due to channel degradation, congestion, queueing effects, or congestion-control behavior~\cite{mccarthy2021congestion,thandavasamy2020cooperative}, the reconstructed collaborator feature can contain missing or stale feature evidence. Our burst-drop setting approximates this failure mode by removing contiguous BEV regions from non-ego collaborator features at inference time.

This stress test therefore evaluates whether the fusion model can preserve detection performance when delayed collaborator evidence is also spatially incomplete. The results in Sec.~4.5 show that the proposed residual interface degrades more gracefully than LRCP~\cite{lrcp} under this joint delay-and-drop setting, supporting the use of conservative residual correction when temporal history is fragmented or unreliable.

\begin{figure}[h!]
    \centering
    
    \begin{minipage}[b]{0.48\linewidth}
        \centering
        \includegraphics[width=\linewidth]{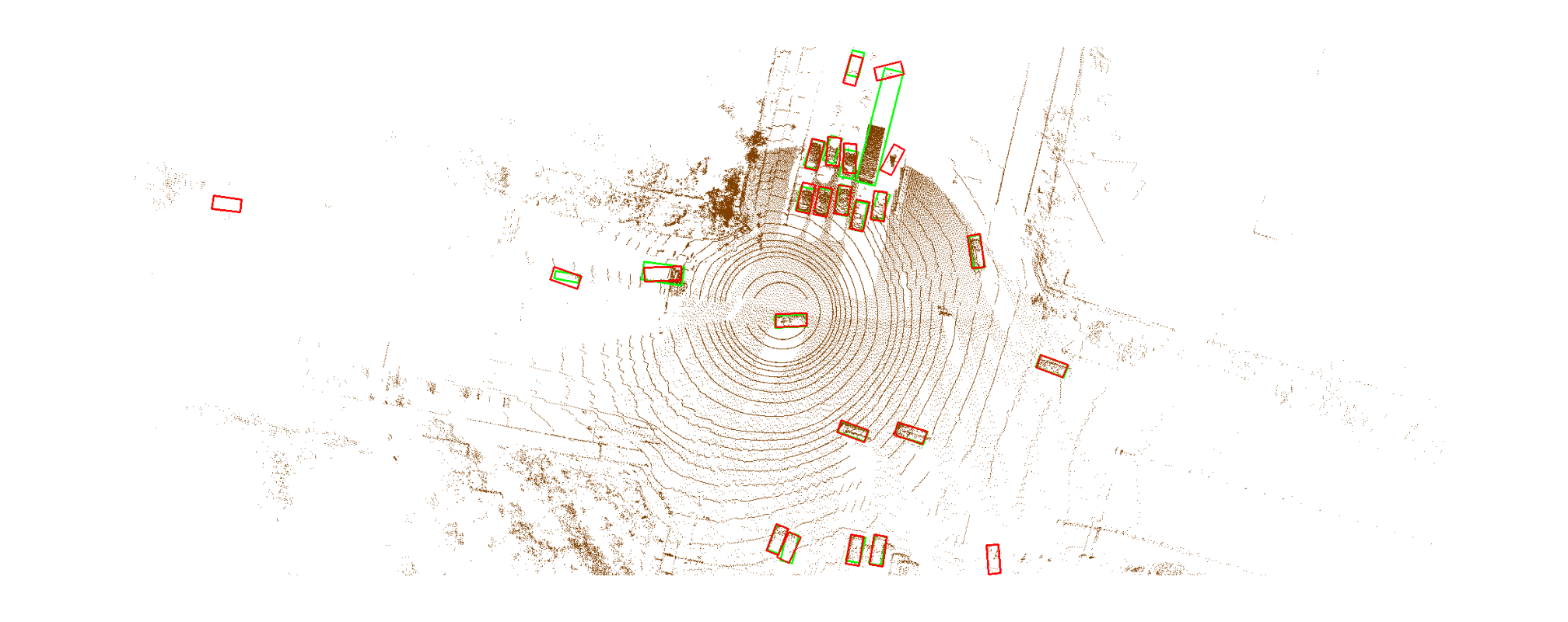}
    \end{minipage}\hfill
    \begin{minipage}[b]{0.48\linewidth}
        \centering
        \includegraphics[width=\linewidth]{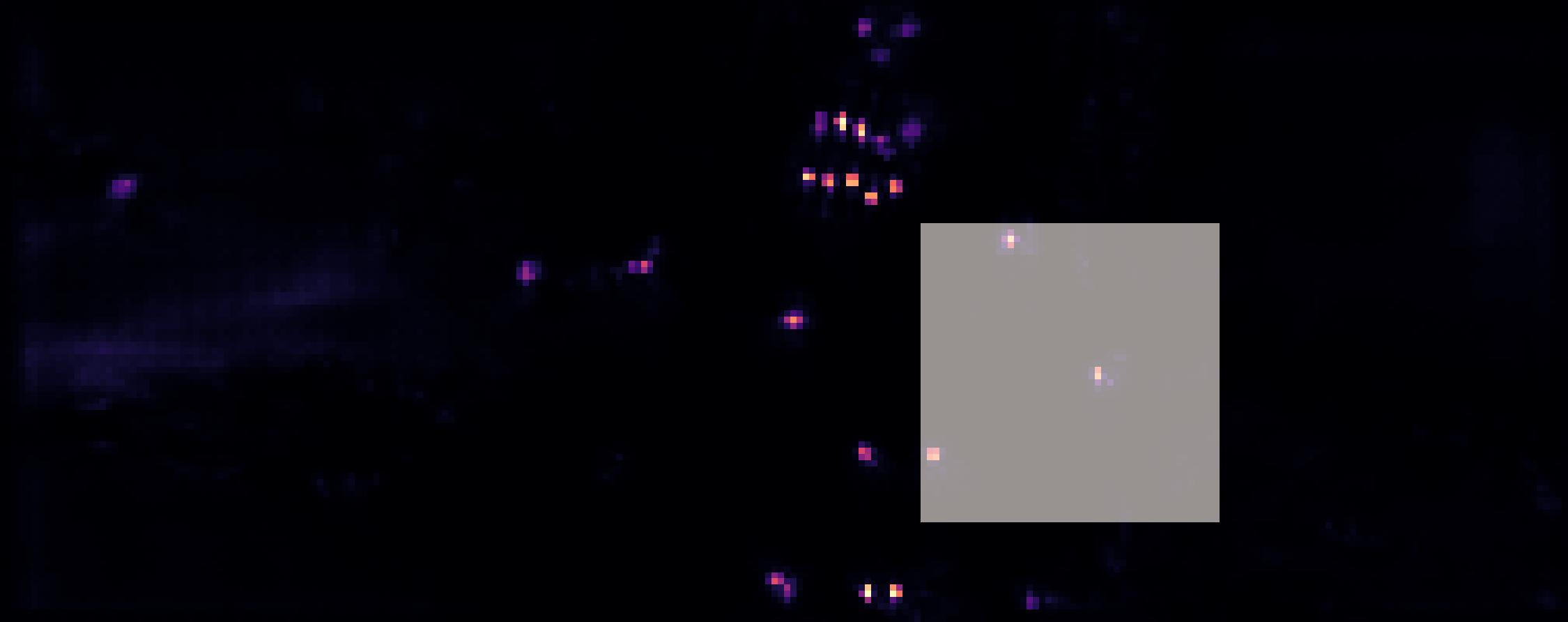}
    \end{minipage}
    
    \vspace{2mm} 
    
    \begin{minipage}[b]{0.48\linewidth}
        \centering
        \includegraphics[width=\linewidth]{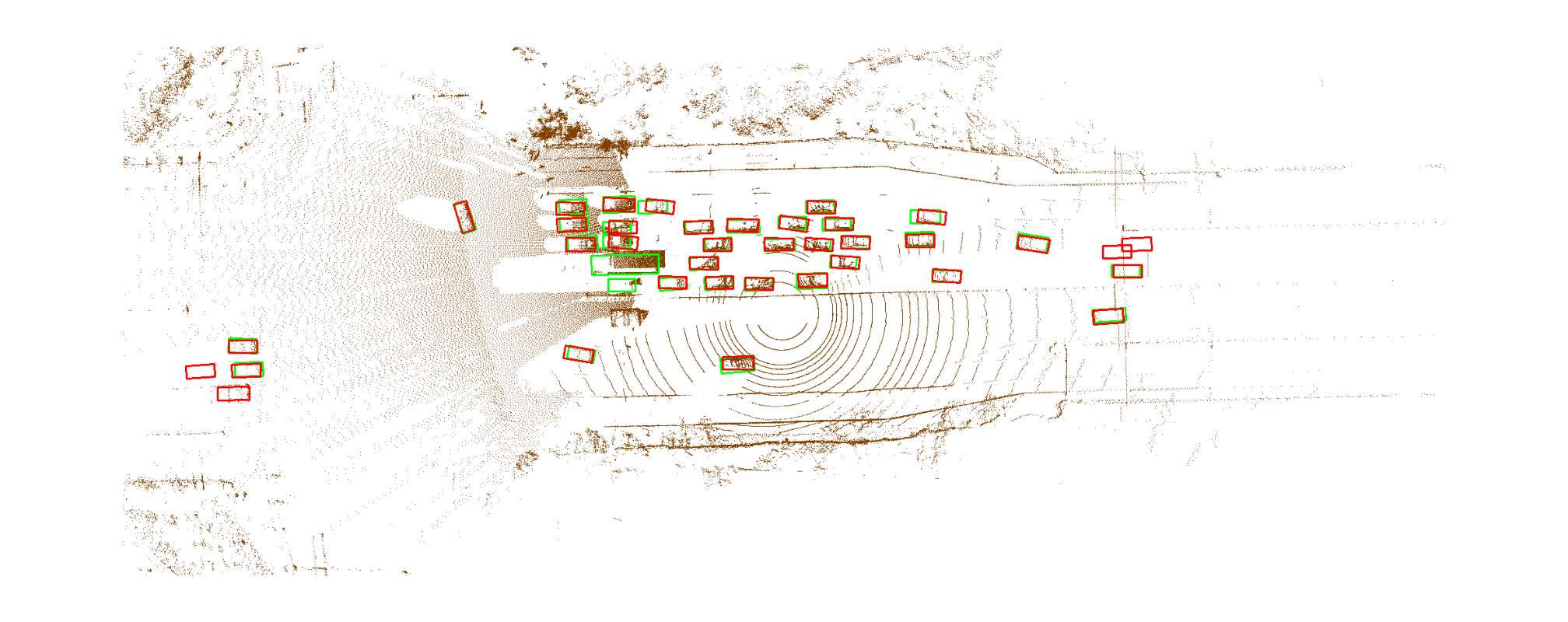}
    \end{minipage}\hfill
    \begin{minipage}[b]{0.48\linewidth}
        \centering
        \includegraphics[width=\linewidth]{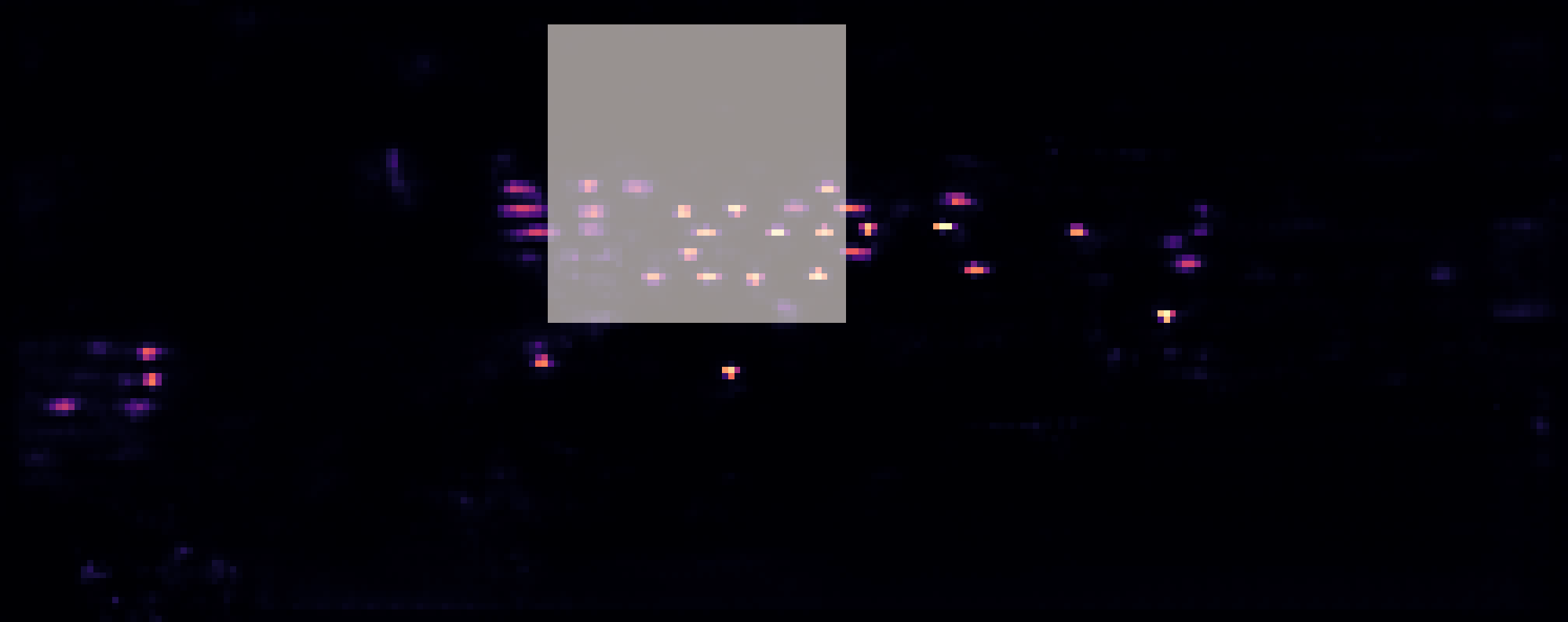}
    \end{minipage}
    
    \caption{\textbf{Burst packet-drop visualization.}
    Top row: First scenario. Bottom row: Second scenario. Left: BEV detection result. Right: detector confidence map with burst-dropped BEV cells highlighted in gray.}
    \label{fig:burst_packet_drop_vis}
\end{figure}

Figure~\ref{fig:burst_packet_drop_vis} visualizes the packet-drop stress used in our robustness experiments. The left panel shows the standard BEV detection output for a synchronized sample. The right panel shows the detector confidence map for the same sample when burst packet drop is applied to the non-ego collaborator feature. Gray regions indicate contiguous BEV cells removed before fusion. This setting is intended as a severe missing-evidence stress test rather than a calibrated wireless-channel model.

\section{Qualitative Examples of Detections under Delay and Pose Noise}
\begin{figure}[h!]
    \centering
    
    \begin{minipage}[b]{0.48\linewidth}
        \centering
        \includegraphics[width=\linewidth]{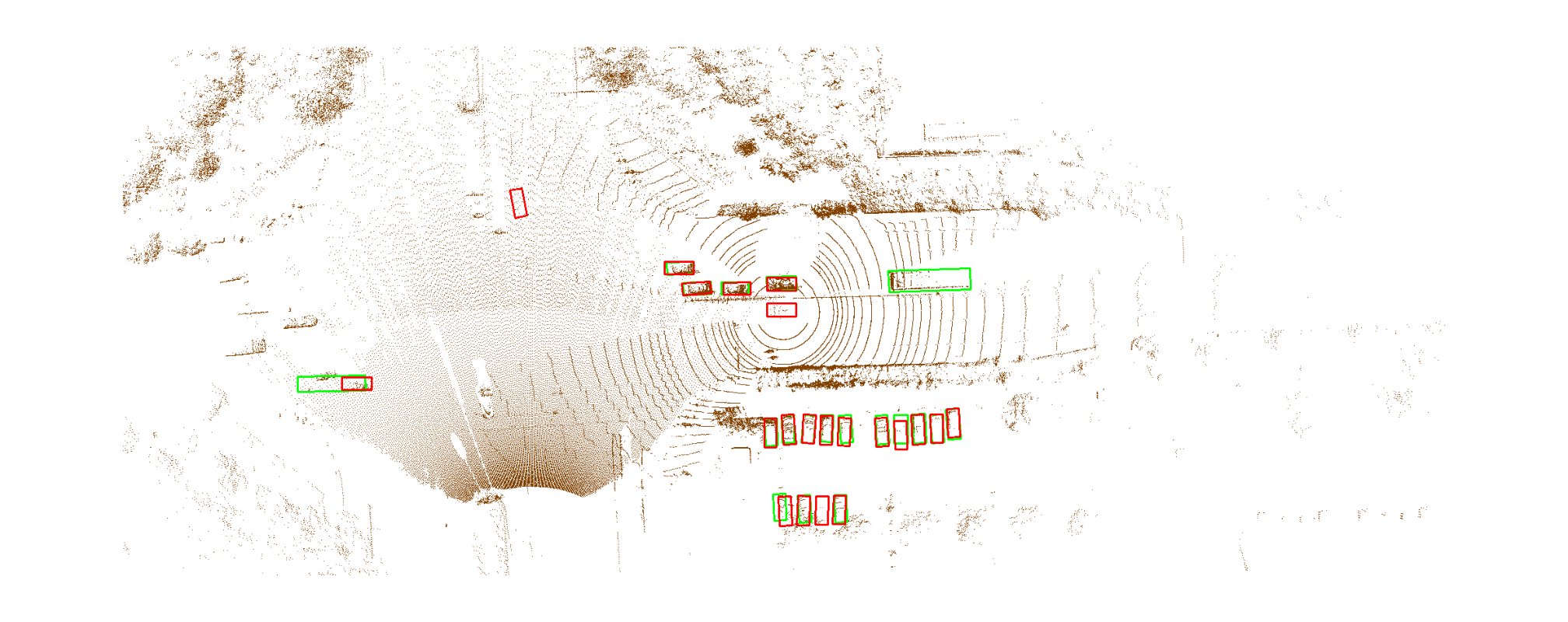}
    \end{minipage}\hfill
    \begin{minipage}[b]{0.48\linewidth}
        \centering
        \includegraphics[width=\linewidth]{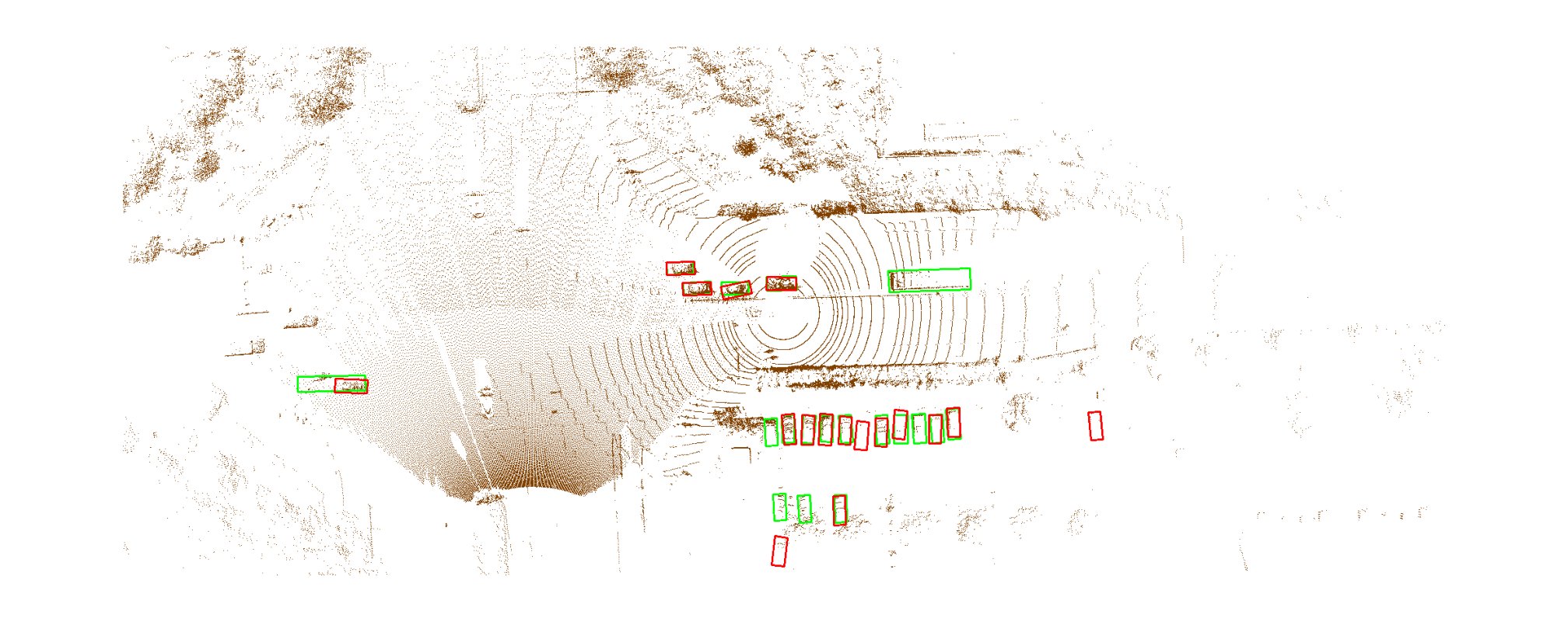}
    \end{minipage}
    
    \vspace{1mm}
    
    \begin{minipage}[b]{0.48\linewidth}
        \centering
        \includegraphics[width=\linewidth]{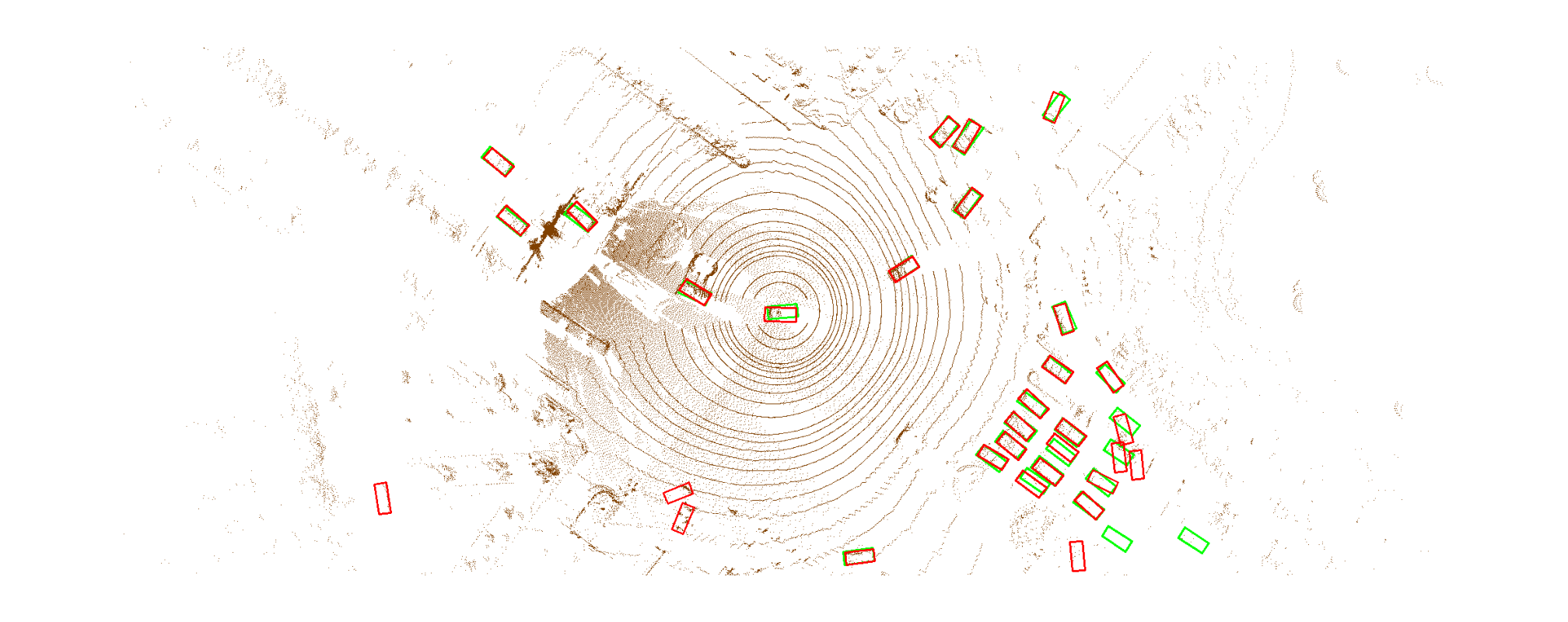}
    \end{minipage}\hfill
    \begin{minipage}[b]{0.48\linewidth}
        \centering
        \includegraphics[width=\linewidth]{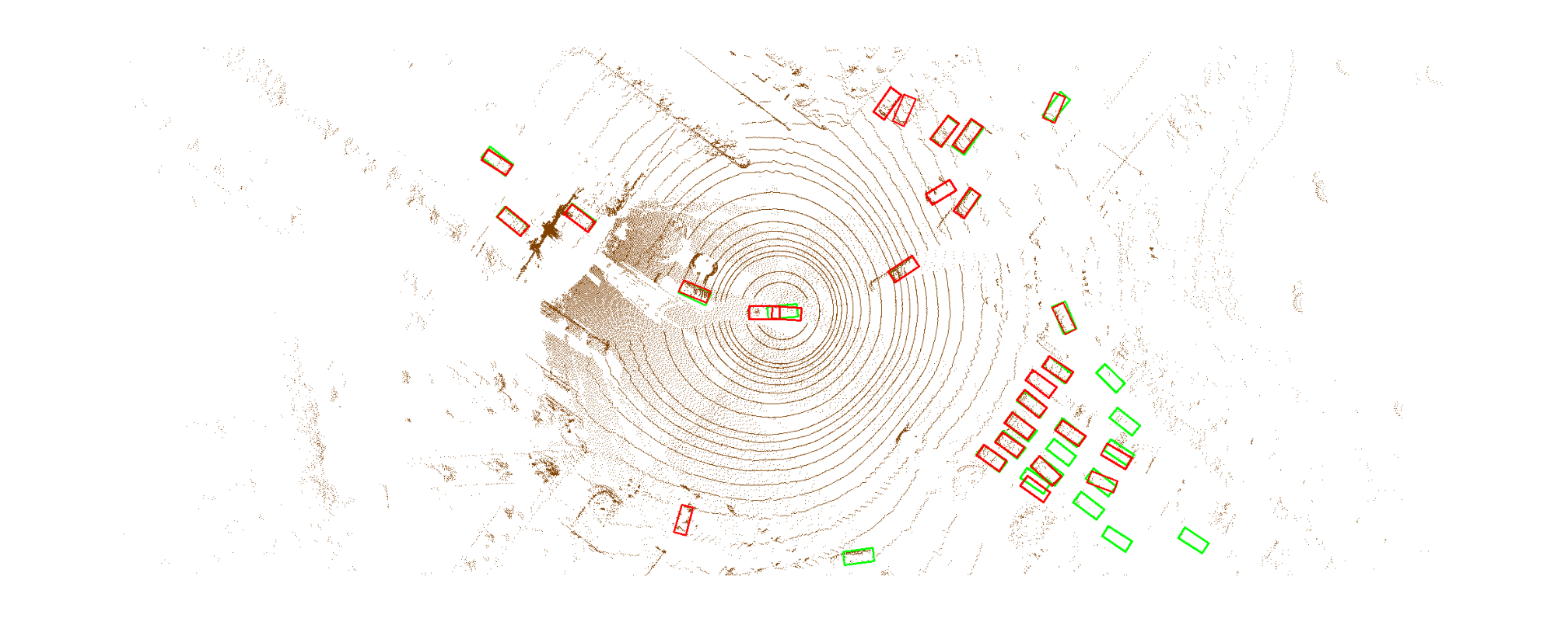}
    \end{minipage}
    
    \caption{\textbf{Qualitative BEV detections under delay and pose noise.}
    Left: ours. Right: LRCP. Top: irregular 500\,ms delay. Bottom: irregular 500\,ms delay with pose noise (0.3\,m / 0.3$^\circ$).}
    \label{fig:bev_detections}
\end{figure}

Figure~\ref{fig:bev_detections} shows representative DAIR-V2X qualitative examples comparing our method with LRCP. The top row shows an irregular 500\,ms delay case, and the bottom row shows irregular 500\,ms delay with additional pose noise (0.3\,m / 0.3$^\circ$). These examples are not intended to show failure-free detection. Both methods can produce false positives under severe temporal misalignment. The main qualitative difference is the structure of the errors: LRCP often introduces duplicated or spatially shifted boxes in dense vehicle regions, consistent with unreliable direct transport under large delay, while our residual temporal fusion tends to keep more detections anchored near the visible object support. This behavior aligns with the quantitative AP@0.7 trend, where our method preserves localization better under severe delay and perturbation despite remaining imperfect.
\section{Runtime Analysis}

\begin{figure}[htbp]
    \centering
    \includegraphics[width=0.8\linewidth, keepaspectratio]{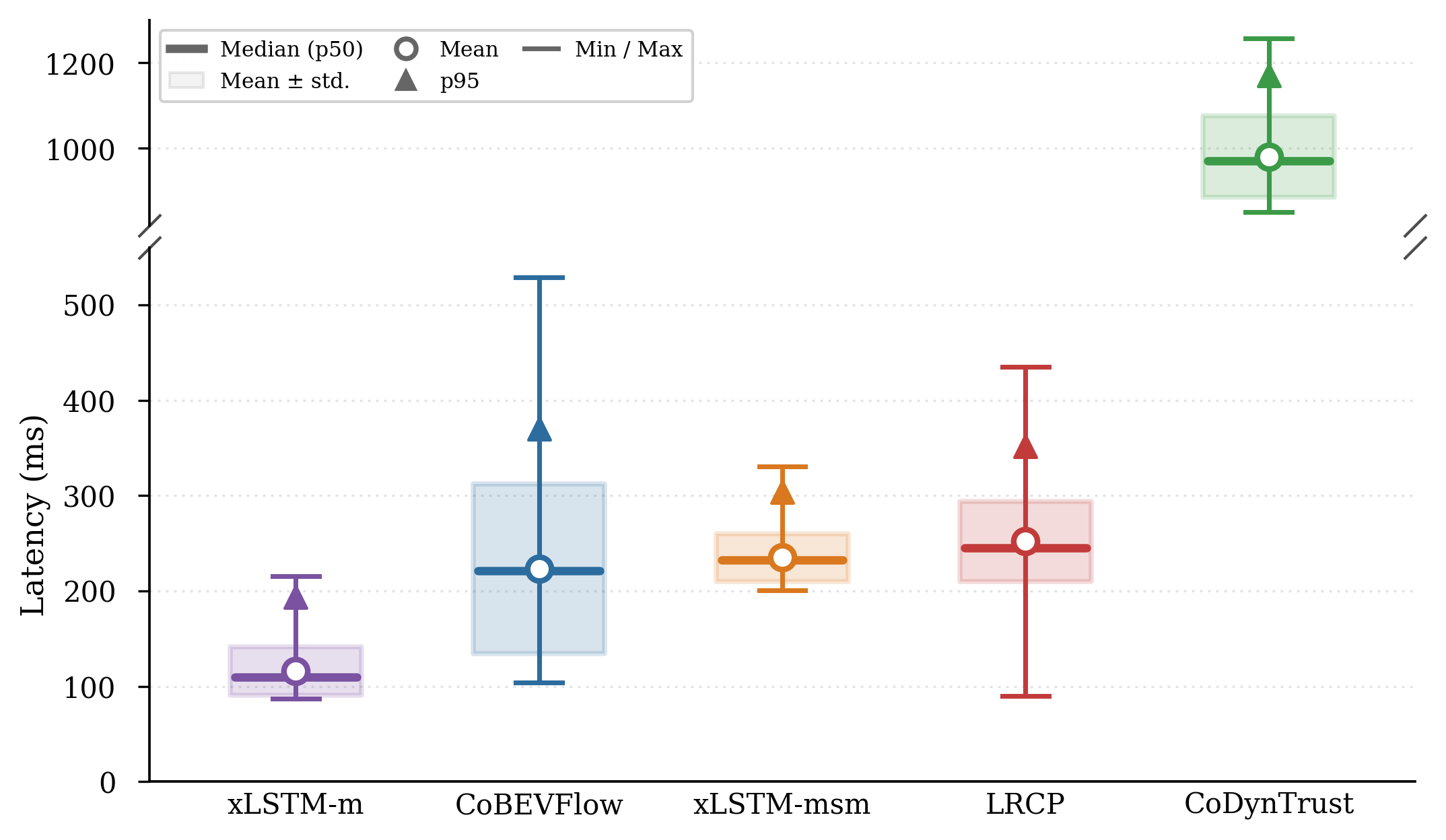}
    \caption{\textbf{Latency comparison.} Runtime measured on RTX 3090.}
    \label{fig:runtime_latency}
\end{figure}

Figure~\ref{fig:runtime_latency} reports inference runtime per sample measured on an RTX 3090. It is measured during inference with batch size 1, identical BEV resolution,
the same number of collaborating agents, 50 warm-up iterations, 200 measured
iterations, and CUDA synchronization before and after each measurement. The lightweight xLSTM-$m$ variant is substantially faster than flow-guided baselines, while the final xLSTM-$m$-$s$-$m$ configuration has runtime comparable to LRCP and much lower than CoDynTrust. This indicates that the residual temporal interface adds moderate recurrent overhead but avoids the high cost of trust-filtered dense compensation.

\section{CoBEVFlow reproduction diagnostic.}
A similar degradation trend for CoBEVFlow on DAIR-V2X is also reported by LRCP~\cite{lrcp}, suggesting that dense flow-based compensation is sensitive to correspondence and masking choices in real-world V2I scenes. We further audited the reproduced CoBEVFlow integration because its in-framework result remains lower than the published reference number. The released CoBEVFlow code path applies the object ROI as a hard multiplicative mask after BEV-flow warping, which suppresses all non-ROI background features. This is a potentially important implementation detail because BEV fusion still relies on contextual and static-scene evidence outside object boxes. To isolate this effect, we evaluated four inference-time variants at the standard DAIR-V2X $p=0.3$ setting: the released hard-mask path, removing the ROI mask, disabling the warp with no mask, and a background-preserving ROI blend that applies the warped feature inside the ROI while retaining the original feature elsewhere.

\begin{table}[h!]
\centering
\caption{\textbf{CoBEVFlow ROI-mask diagnostic.} DAIR-V2X AP at $p=0.3$ for reproduced CoBEVFlow. The background-preserving blend matches the method description more closely than the released hard-mask multiplication.}
\label{tab:cobevflow_roi_diag}
\small
\setlength{\tabcolsep}{6pt}
\begin{tabular}{lccc}
\toprule
Variant & AP@0.3 & AP@0.5 & AP@0.7 \\
\midrule
Released hard ROI mask & 0.810 & 0.739 & 0.597 \\
No ROI mask & 0.822 & 0.760 & 0.611 \\
Identity warp, no mask & 0.822 & 0.760 & 0.611 \\
Background-preserving ROI blend & 0.822 & 0.760 & 0.611 \\
\bottomrule
\end{tabular}
\end{table}

Table~\ref{tab:cobevflow_roi_diag} shows that the hard ROI mask is the main source of degradation in our reproduction. Replacing it with the background-preserving blend improves AP@0.7 from 0.597 to 0.611, matching the no-mask and identity-warp controls. However, the flow warp itself does not improve over identity in this setting. We therefore keep the released hard-mask path as the reproduced CoBEVFlow baseline in Table~1 in main paper, and use the remaining rows only as diagnostics explaining why this reproduced baseline is lower than the published reference.

\section{Synchronized/asynchronous training trade-off}
To test whether robustness under severe delay inherently requires sacrificing
synchronized accuracy, we additionally fine-tune the reported checkpoint at
low learning rate using 10\% synchronized and 90\% causal-asynchronous
samples. The resulting checkpoint reaches 0.669/0.648/0.641 AP@0.7 under
Sync/Irr.\ 300\,ms/Irr.\ 500\,ms, compared with
0.664/0.640/0.635 for the reported model at Table 1 in the main paper. It therefore matches LRCP under
synchronized input while preserving the delayed advantage, indicating that
the synchronized gap of the reported checkpoint is primarily a
training-distribution trade-off rather than an intrinsic limitation of the
residual architecture.
\bibliographystyle{splncs04}
\bibliography{main}
\end{document}